\documentclass[letterpaper, 10 pt, conference]{ieeeconf}  

\IEEEoverridecommandlockouts                              

\usepackage[table,xcdraw]{xcolor}
\definecolor{awesome}{rgb}{1.0, 0.13, 0.32}

\usepackage{adjustbox}
\usepackage{graphics} 
\usepackage{epsfig} 
\usepackage[color=red]{todonotes}
\usepackage{mathptmx} 
\usepackage{times} 
\usepackage{amsmath} 
\usepackage{amssymb}  
\usepackage{microtype}
\usepackage{multirow}
\usepackage{vcell}
\usepackage{svg}
\usepackage{cite}

\makeatletter
\let\NAT@parse\undefined
\makeatother
\usepackage[colorlinks=true,linkcolor=red,citecolor=blue]{hyperref}

\title{\LARGE \bf
MaskSem: Semantic-Guided Masking for Learning 3D Hybrid High-Order Motion Representation
}

\author{Wei Wei, Shaojie Zhang, Yonghao Dang, Jianqin Yin$^{*}$~\IEEEmembership{}
\thanks{*Corresponding author. Jianqin Yin}
\thanks{Wei Wei, Shaojie Zhang, Yonghao Dang, Jianqin Yin are with the School of Intelligent Engineering and Automation, Beijing University of Posts and Telecommunications, Beijing 100876, China (e-mail:ww2024@bupt.edu.cn; zsj@bupt.edu.cn; dyh2018@bupt.edu.cn; jqyin@bupt.edu.cn)}}

\begin{document}

\maketitle
\thispagestyle{empty}
\pagestyle{empty}

\begin{abstract}
Human action recognition is a crucial task for intelligent robotics, particularly within the context of human-robot collaboration research. In self-supervised skeleton-based action recognition, the mask-based reconstruction paradigm learns the spatial structure and motion patterns of the skeleton by masking joints and reconstructing the target from unlabeled data. However, existing methods focus on a limited set of joints and low-order motion patterns, limiting the model's ability to understand complex motion patterns. To address this issue, we introduce MaskSem, a novel semantic-guided masking method for learning 3D hybrid high-order motion representations. This novel framework leverages Grad-CAM based on relative motion to guide the masking of joints, which can be represented as the most semantically rich temporal orgions. The semantic-guided masking process can encourage the model to explore more discriminative features. Furthermore, we propose using hybrid high-order motion as the reconstruction target, enabling the model to learn multi-order motion patterns. Specifically, low-order motion velocity and high-order motion acceleration are used together as the reconstruction target. This approach offers a more comprehensive description of the dynamic motion process, enhancing the model's understanding of motion patterns. Experiments on the NTU60, NTU120, and PKU-MMD datasets show that MaskSem, combined with a vanilla transformer, improves skeleton-based action recognition, making it more suitable for applications in human-robot interaction. The source code of our MaskSem is available at \textit{\href{https://github.com/JayEason66/MaskSem}{https://github.com/JayEason66/MaskSem}}.

\end{abstract}

\section{INTRODUCTION}
Skeleton-based action recognition plays a crucial role in enabling robots to understand and predict human actions, which is an essential capability for tasks such as human-robot interaction and collaborative robotics \cite{ wen2023interactive, xing2022understanding}. Skeleton-based action recognition \cite{wang2023comprehensive} leverages the positional information of human joints, making it more robust to environmental factors such as lighting and occlusions. This approach has become a more reliable solution for action recognition in robotic systems, offering improved resilience while accurately modeling human motion. Focusing on skeletal data \cite{yan2018spatial, chen2021channel, plizzari2021spatial} enables robots to understand human activities better and respond accordingly in dynamic environments.

Although supervised skeleton-based action recognition methods \cite{shi2019two, si2019attention} achieve impressive performance, their reliance on expensive and labor-intensive annotations ultimately limits their generalization ability and hinders the recognition of subtle actions. To address these challenges, self-supervised learning methods \cite{wu2023skeletonmae, mao2023masked, yan2023skeletonmae, lin2023actionlet} have emerged as a promising research direction in the field of skeleton-based action recognition. Unlike traditional supervised methods, self-supervised learning utilizes the inherent structure of the data through carefully designed pretext tasks, enabling the model to learn more effectively from the data. Recently, masked autoencoders (MAE) \cite{he2022masked} have been employed to utilize masked self-reconstruction as a pretraining task for skeleton-based action recognition. In these approaches, the input skeleton sequence is deliberately masked or corrupted. The model is trained to reconstruct the original sequence, enabling the autoencoder to learn essential features and develop compact, informative representations by inferring various reconstruction targets (e.g., masked joint coordinates \cite{wu2023skeletonmae,yan2023skeletonmae}, temporal motion \cite{mao2023masked}, or joints feature \cite{zhu2024motion}).

\begin{figure}[t!]
    \centering
    \includegraphics[width=1\linewidth,trim=100 50 140 60,clip]{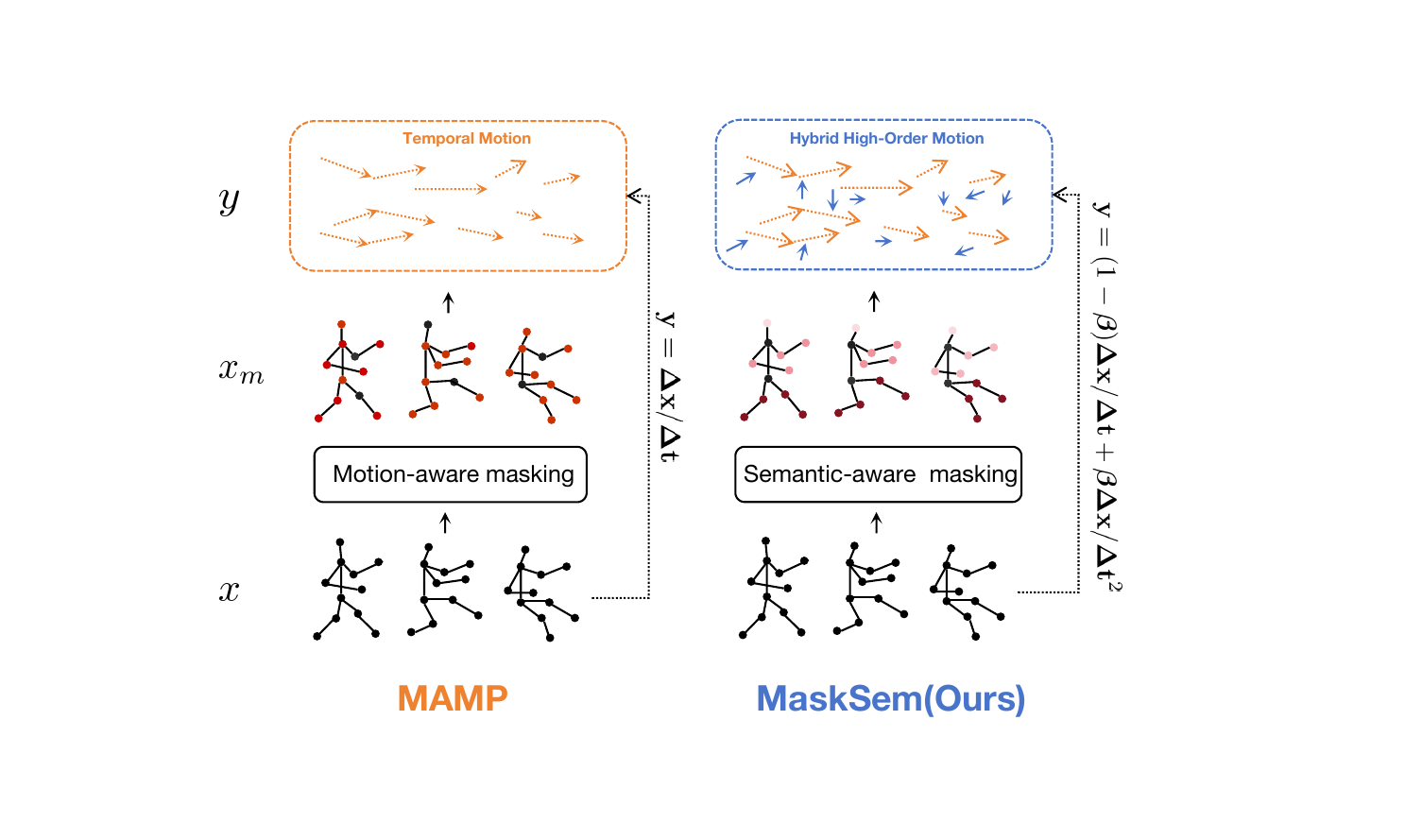}
    \caption{A comparison of the masking strategy and reconstruction targets between masked motion prediction (MAMP) and our semantic-guided masking hybrid high-order motion prediction (MaskSem).}
    \label{fig1}
\end{figure}

 However, most existing MAE-based models exhibit notable shortcomings. First, regarding the masking strategy, \textbf{random masking \cite{wu2023skeletonmae, yan2023skeletonmae} fails to rely on the inherent structure of the data and does not utilize the contextual information from the input data or the learning feedback from the model.} This limits the model's ability to learn discriminative features (such as joint positions, angles, and motion details), thereby impairing its capability to distinguish between different action categories. \textbf{Binary masking \cite{mao2023masked, zhu2024motion} overlook the contributions of different joints and fail to fully leverage semantic information to learn discriminative features.} This prevents the model from distinguishing joint importance, weakening its understanding of motion patterns and action discrimination. \textbf{Secondly, the reconstruction targets of existing methods \cite{mao2023masked} focus solely on low-order motion patterns.} This makes it difficult to effectively capture subtle variations in motion (e.g., reading vs. writing). Motivated by these observations, we pose the following question: \textit{“How can we design a masking strategy with contextual information and multi-order motion patterns to enhance learning? ”}

This work proposes a novel training method to address the issues above, as illustrated in Fig. \ref{fig1}. In most action sequences, the background regions remain stationary, while motion is typically confined to localized areas (e.g., hands or head). Therefore, we use Grad-CAM \cite{selvaraju2020grad} relative to the average motion as semantic information to mask the most important joints in the input skeleton sequence, assigning weights to each masked joint based on its contributions for computing the reconstruction loss. This masking strategy enables more flexible and adaptive capture of the features of discriminative joints, dynamically adjusting the masking process based on each skeleton joint’s contribution to the action classes. Afterward, the unmasked portion is fed into the encoder for feature extraction, generating the latent representation of the skeleton sequence. Finally, the encoder output is passed to the decoder, where we perform hybrid high-order motion reconstruction of the input skeleton sequence. Hybrid high-order motion includes the first-order and second-order differences of the input joint coordinates, which we refer to as the velocity and acceleration of the joints. Velocity reveals the dynamic motion of the joint. Acceleration represents the change in the joint's motion rate, indicating whether there are rapid changes during the motion. This reconstruction of multi-order motion patterns provides a more comprehensive description of the dynamic motion process, enhances the model's understanding of motion variations, and effectively captures subtle changes in motion.

Overall, we make the following three key contributions:
\begin{enumerate}
    \item We propose an innovative semantic-guided masking strategy that dynamically adjusts the masking process based on the importance of joints, aiming to guide the model to prioritize joints that contribute more to discriminative information. 
    \item We present for the first time masked hybrid motion prediction to learn high-order 3D action representation, allowing the pre-trained model to acquire an adequate understanding of the complex motion patterns within spatially sparse but temporally dense skeleton sequences.
    \item We conduct experiments on three widely used benchmarks to demonstrate the effectiveness of our approach. Our proposed MaskSem method achieves competitive performance across these datasets, showcasing its potential for improving skeleton-based action recognition.
\end{enumerate}

\section{RELATED WORK}
\subsection{Skeleton-Based Action Recognition}
Skeleton-based action recognition has advanced rapidly in computer vision. Early methods relied on handcrafted features to capture geometric relationships between joints \cite{ohn2013joint}, but struggled with dynamic movements and temporal dependencies. Deep learning models, such as RNNs, hierarchical RNNs, and Spatio-Temporal LSTMs, addressed temporal learning but faced issues like vanishing gradients \cite{liu2017skeleton,liu2016spatio}. Graph Convolutional Networks (GCNs), exemplified by ST-GCN, enabled effective modeling of spatial and temporal dynamics \cite{kipf2016semi,yan2018spatial}. Attention mechanisms further enhanced GCNs by focusing on relevant joints or frames \cite{chen2021channel,chi2022infogcn}. More recently, Transformer architectures have been widely adopted for capturing complex spatiotemporal interactions in skeleton sequences \cite{vaswani2017attention,plizzari2021skeleton}.
\subsection{Masked Skeleton Modeling}
Masked autoencoders (MAE) draw inspiration from masked language modeling in NLP (e.g., BERT \cite{devlin2018bert}), 
extending the idea of randomly masking input tokens and predicting the missing content to the visual domain. 
He et al. \cite{he2022masked} first formalized this concept for image representation learning. Building on it, 
SkeletonMAE \cite{wu2023skeletonmae} applied MAE to skeleton-based action recognition by predicting the original 
coordinates of masked joints, showing the potential of self-supervised learning to leverage unlabeled skeleton data. 
MAMP \cite{mao2023masked} then specialized in reconstructing temporal motion for masked skeleton regions, emphasizing 
the importance of dynamic motion cues. MMFR \cite{zhu2024motion} combined natural language supervision with 
masked skeleton features, harnessing the semantic richness of language models. S-JEPA \cite{abdelfattah2024s} predicts high-level features from partially masked skeleton sequences.

However, most existing 3D action representation methods still rely on binary masking strategies, which ignore the 
relative importance of individual joints and connections and thus fail to highlight critical regions essential 
for understanding actions. To address this limitation, we propose a semantic-guided 
masking strategy that assigns weights to joints.

\section{METHODOLOGY}
\subsection{Pipeline Overview}
\begin{figure*}[ht]
    \centering
    \includegraphics[height=8cm,trim=110 205 215 50, clip]{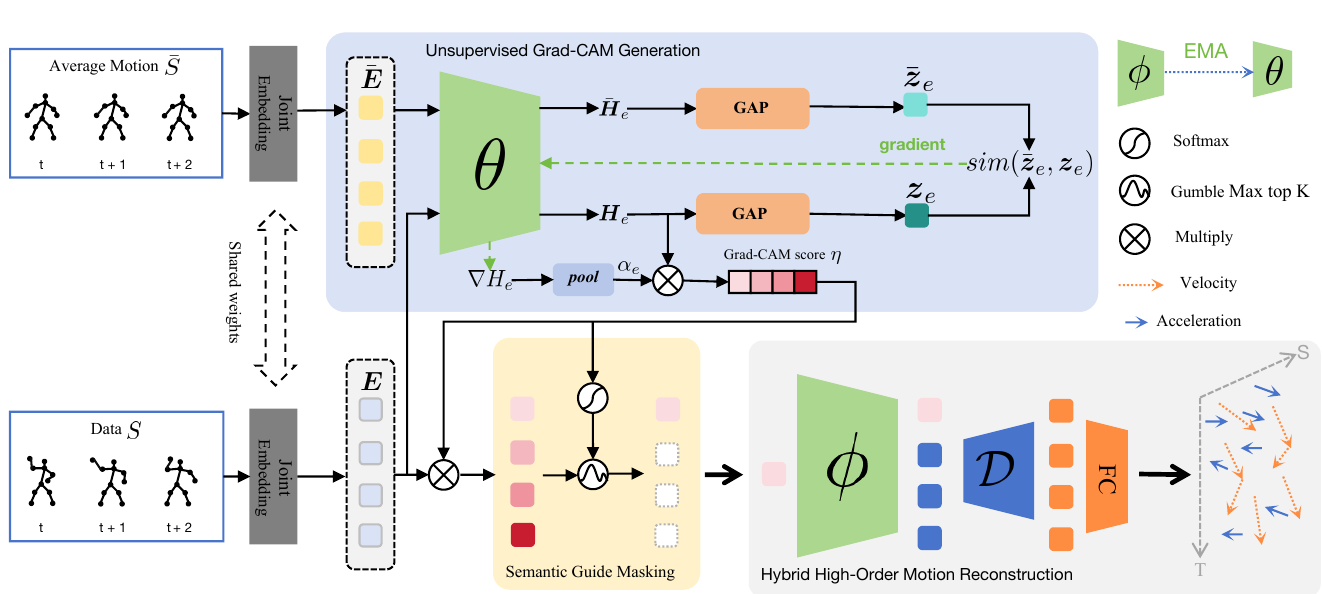}
    \caption{\textbf{Architecture Overview of MaskSem.} MaskSem is primarily consists of three components: Unsupervised Grad-CAM Generation, Semantic Guided Masking, and Hybrid High-Order Motion Reconstruction. The pre-trained encoder $\boldsymbol{\phi}$ can be utilized independently in downstream discriminative tasks.}
    \label{fig2}
\end{figure*}
Figure~\ref{fig2} illustrates the process of our semantic-guided masking and hybrid high-order motion prediction framework. The input skeleton sequence first is passed through a Joint Embedding module to reduce temporal redundancy, enhancing computational efficiency. Spatial and temporal positional embeddings are then added to improve the model's ability to capture precise positional information. The features produced by the Encoder $\boldsymbol{\theta}$ are utilized to compute the Grad-CAM of relative motion, which serves as semantic information to guide the masking process. Finally, in the hybrid high-order motion reconstruction module, the unmasked joints are fed into the Encoder $\boldsymbol{\phi}$, and then concatenated with the masked joints before being passed into the Decoder $D$ to reconstruct the hybrid high-order motion of the masked joints. To guide learning, the mean squared error (MSE) loss is computed between the reconstructed features and the hybrid high-order motion information from the original sequence.

\subsection{Joint Embedding} \label{subsec:joint_embedding}
The input to our method is a skeleton sequence $\boldsymbol{S} \in \mathbb{R}^{T \times V \times C}$, where $T$ is the fixed temporal length, $V$ represents the number of joints, and $C$ is the number of coordinate channels. The sequence is obtained by randomly cropping the original data and resizing it to the desired temporal length.

To address temporal redundancy in skeleton sequences, we follow the approach in \cite{mao2023masked} and divide the input sequence $\boldsymbol{S}$ into $l$ equally non-overlapping segments $\boldsymbol{S}' \in \mathbb{R}^{T_e \times V \times l \times C}$, where $T_e = T / l$ is the length of each segment, and $l$ is the number of segments. 
\begin{equation}
\boldsymbol{E}' = Joint Embedding(\boldsymbol{S}') \in \mathbb{R}^{T_e \times V  \times C_e}
\label{equ-1}
\end{equation}

\subsection{Unsupervised Grad-CAM Generation} \label{subsec:Grad-CAM-Generation}
Grad-CAM \cite{selvaraju2020grad} is a widely used technique for saliency detection that generates an attention map by computing the gradient of the target 
 loss with respect to the intermediate features of the network. This map highlights the regions with the most influential features in reducing the loss. In the context of our proposed self-supervised approach, no labels are required during the training process.

\noindent \textbf{Average Motion as a Static Reference.} Inspired by ActCLR \cite{lin2023actionlet}, we calculate the similarity between the action sequence and a static sequence, which we assume to have no motion, to identify the important joints in the skeletal features. Mathematically, this can be expressed as follows:
\begin{equation}
\bar{\boldsymbol{S}} = \frac{1}{N} \sum_{i=1}^{N} \boldsymbol{S}_i
\label{equ-4}
\end{equation}
where $\boldsymbol{S}_i$ represents the $i\text{-th}$ skeleton sequence, and $N$ is the total number of sequences in the dataset.\\
\textbf{Grad-CAM Generation.} We first apply local pooling to the average motion $\bar{\boldsymbol{S}} $ to obtain $\bar{\boldsymbol{S}'} $, which is then passed into the Joint Embedding module. Next, spatiotemporal positional encodings are added, and finally, a reshape operation is performed to obtain $\bar{\boldsymbol{E}} $. The detailed process is as follows:
\begin{equation}
\begin{aligned}
    \bar{\boldsymbol{E}'} &= JointEmbedding(\bar{\boldsymbol{S}'}) \in \mathbb{R}^{T_e \times V  \times C_e} \\
    \bar{\boldsymbol{E}_e} &= \bar{\boldsymbol{E}'} + \boldsymbol{P}_t + \boldsymbol{P}_s \\
    \bar{\boldsymbol{E}} &= \mathit{reshape}(\bar{\boldsymbol{E}_e}) \in \mathbb{R}^{N \times C_e}
\end{aligned}
\label{equ-5}
\end{equation}
\hspace*{1em}Next, the processed tokens are fed into the offline Encoder ${\boldsymbol{\theta}}$ to obtain the corresponding dense features. The offline Encoder ${\boldsymbol{\theta}}$ extracts latent representations using the vanilla transformer \cite{vaswani2017attention} architecture with $L_e$ layers. 
\begin{equation}
\begin{aligned}
    \bar{\boldsymbol{H}}_0 & = \bar{\boldsymbol{E}}, \\
    \bar{\boldsymbol{H}}'_l & = \text{MSA}(\text{LN}(\bar{\boldsymbol{H}}_{l-1})) + \bar{\boldsymbol{H}}_{l-1}, \quad l \in 1, \cdots, L_e, \\
    \bar{\boldsymbol{H}}_l & = \text{MLP}(\text{LN}(\bar{\boldsymbol{H}}'_l)) + \bar{\boldsymbol{H}}'_l, \quad l \in 1, \cdots, L_e, \\
    \bar{\boldsymbol{H}}_e & = \text{LN}(\bar{\boldsymbol{H}}_{L_e}).
\end{aligned}
\label{equ-6}
\end{equation}
where $\bar{\boldsymbol{H}}_e$ represents the dense features obtained from the average motion. The same operation is applied to the input sequence to obtain the corresponding dense features, denoted as $\boldsymbol{H}_e$.

We apply global average pooling (GAP) on both $\boldsymbol{H}_e$ and $\bar{\boldsymbol{H}}_e$ along the temporal and spatial dimensions to obtain the global features $\boldsymbol{z}_e$ and $\bar{\boldsymbol{z}}_e$, respectively. Then, we compute the cosine similarity between these two global features:
\begin{equation}
\begin{aligned}
    \text{sim}(\boldsymbol{z}_e, \bar{\boldsymbol{z}}_e) & = \frac{\boldsymbol{z}_e \cdot \bar{\boldsymbol{z}}_e}{\|\boldsymbol{z}_e\| \| \bar{\boldsymbol{z}}_e \|}.
\end{aligned}
\label{equ-8}
\end{equation}
\hspace*{1em}To identify regions with significantly larger motion amplitudes relative to the average, we take the gradient of the inverse similarity and backpropagate it, mapping it onto the dense feature $\boldsymbol{H}_e$. Then, by performing average pooling across the temporal and spatial dimensions, we obtain the neuron importance weights $\alpha_e$.
\begin{equation}
\alpha_e = \frac{1}{T \times V} \sum_{t=1}^{T} \sum_{v=1}^{V} \textbf{ReLU}\left( \frac{\partial (-\text{sim}(\boldsymbol{z}_e, \bar{\boldsymbol{z}_e}))}{\partial \boldsymbol{H}_e} \right)
\label{equ-9}
\end{equation}
\hspace*{1em}These importance weights reflect the influence of each channel dimension on the resulting difference. Consequently, these weights $\alpha_e$ are treated as the difference activation map. The Grad-CAM score vector $\mathbf{\eta} = \{\eta_i\}_{i=1}^n \in \mathbb{R}^n$ is computed using $\alpha_e$ and $\boldsymbol{H}_e$. To obtain the final result, we compute a weighted combination of the difference activation map and the dense features, expressed as:
\begin{equation}
\mathbf{\eta} = \textbf{ReLU} \left( \sum_{c=1}^{C} \alpha_{e} \boldsymbol{H}_{e} \right) A_{vv}
\label{equ-10}
\end{equation}
where, $A_{vv}$ denotes the adjacency matrix of the skeleton data used for importance smoothing. The application of $\text{ReLU}(\cdot)$ ensures that the values remain non-negative. Note that the original Grad-CAM calculates the gradients of the feature map from the last transformer layer. Higher Grad-CAM scores in $\eta$ indicate that the corresponding dimension contributes more to data discrimination by reducing similarity.

\subsection{Semantic-Guided Masking} \label{subsec:Grad-CAM-Masking}
The activation scores derived from Grad-CAM provide a task-specific saliency map that highlights the discriminative joints that contribute most to the predictions of the model. Unlike conventional masking strategies, Grad-CAM leverages gradient information to capture the relative importance of each joint. Our goal is to mask the most highly activated joints to reduce the dependency of the model on them and encourage the exploration of a wider range of informative joints.

To utilize Grad-CAM for masking, we first compute the Grad-CAM activation score $\eta$ for all joints. These scores are normalized into a probability distribution $\pi$ using a softmax function:
\begin{equation}
\pi = \textit{softmax}\left( \frac{\eta}{\tau_{grad}} \right) 
\label{equ-11}
\end{equation}
\hspace*{1em}The temperature hyperparameter $\tau_{grad}$ controls the sharpness of the distribution. As illustrated in Fig. \ref{fig3}, with this distribution, we can probabilistically mask joints based on their significance, with joints having higher probabilities being more likely to be masked. This approach allows us to focus on the most activated joints while simultaneously encouraging the model to explore less critical joints, ultimately promoting a more diverse and robust feature representation. The masking is performed using the Gumbel-max trick \cite{gumbel1954statistical}, which enables the sampling of joints in accordance with their activation probabilities:
\begin{equation}
\begin{aligned}
    K & = \delta \times T_e \times V, \\
    r & = -\log(-\log \varepsilon), \quad \varepsilon \in U[0, 1]^{T_e \times V}, \\
    \textit{idx}^{\text{mask}} & = \textit{Index-of-Top-K}(\log \pi + r).
\end{aligned}
\label{equ-12}
\end{equation}
Where $\delta \in [0, 1]$ represents a predetermined proportion of joints to be masked, and $\varepsilon$ is random noise sampled from a uniform distribution between 0 and 1. The obtained $\textit{idx}^{\text{mask}}$ represents the indices of the masked joints, which are used to select tokens for encoding in the Encoder $\boldsymbol{\phi}$. Through this operation, the more active joints are more likely to be masked, thereby encouraging the model to focus on regions with richer semantic information and learn more discriminative features.

\subsection{Hybrid High-Order Motion Reconstruction}
\noindent $\textbf{\text{Encoder}}$ $\boldsymbol{\phi}$\textbf{:} The Encoder $\boldsymbol{\phi}$ is an online encoder, which is used to extract representations. The Encoder $\boldsymbol{\phi}$ architecture here is the same as the one used in the Encoder $\boldsymbol{\theta}$ of Section~\ref{subsec:Grad-CAM-Generation}. Encoder $\boldsymbol{\theta}$ is updated by incrementally applying momentum-based updates to the parameters of Encoder $\boldsymbol{\phi}$.
\begin{equation}
\begin{alignedat}{2}
    \boldsymbol{H}_e^{um} & = {\boldsymbol{\phi}} (\boldsymbol{E}_e^{um}), \\
    \boldsymbol{\theta} & = m\boldsymbol{\theta} + (1 - m)\boldsymbol{\phi}.
\end{alignedat}
\label{equ-13}
\end{equation}
\textbf{Decoder D:} $\boldsymbol{H}_e^{um}$ represents the unmasked joints' representations. Learnable mask tokens are inserted into positions specified by $\textit{idx}^{\text{mask}}$ to form $\boldsymbol{E}_d \in \mathbb{R}^{T_e \times V \times C_e}$, which is then fed into a transformer decoder consisting of $L_d$ layers for masked reconstruction.
\begin{equation}
\begin{aligned}
    \boldsymbol{E}_0 & = \boldsymbol{E}_d + \boldsymbol{P}_t^d + \boldsymbol{P}_s^d, \\
    \boldsymbol{D}'_l & = \text{MSA}(\text{LN}(\boldsymbol{D}_{l-1})) + \boldsymbol{D}_{l-1}, \quad l \in 1, \cdots, L_d, \\
    \boldsymbol{D}_l & = \text{MLP}(\text{LN}(\boldsymbol{D}'_l)) + \boldsymbol{D}'_l, \quad l \in 1, \cdots, L_d, \\
    \boldsymbol{D}_d & = \text{LN}(\boldsymbol{D}_{L_d}),
\end{aligned}
\label{equ-14}
\end{equation}
where $\boldsymbol{P}_t^d$ and $\boldsymbol{P}_s^d$ represent the positional embeddings for the temporal and spatial dimensions of the transformer decoder, respectively.
\begin{figure}
    \centering
    \includegraphics[width=1\linewidth,trim=120 140 150 80, clip]{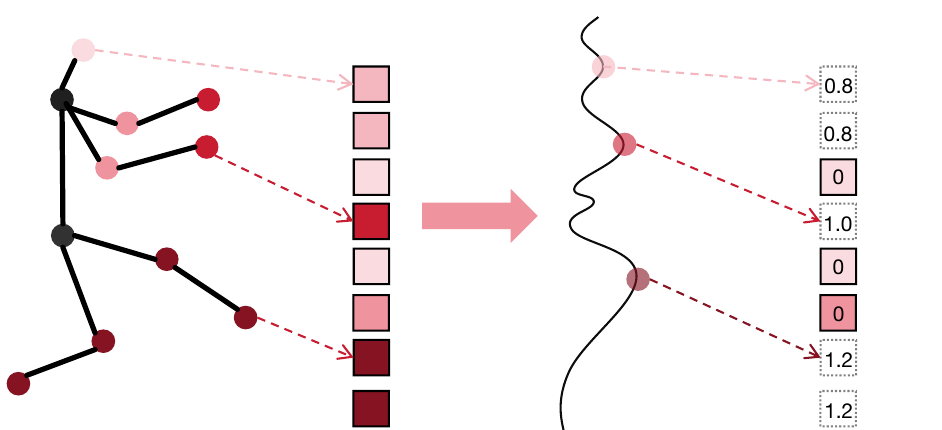}
    \caption{The Grad-CAM scores serve as an empirical semantic richness prior to guiding the masking strategy. Deeper colors signify greater joint importance, increased probabilities of being masked, and more substantial contributions to the loss.}
    \label{fig3}
\end{figure}

\noindent \textbf{Reconstruction Target:} To ensure effective learning, the reconstruction target is not the original skeleton data but rather the hybrid high-order motion of the skeleton, specifically its velocity and acceleration across the temporal dimension. Let the skeleton sequence at time $t$ be denoted as $\boldsymbol{S}_t$. The first-order and second-order temporal derivatives, corresponding to velocity and acceleration, are computed as follows:
\begin{equation}
\begin{aligned}
    \boldsymbol{M}_i^v & = \boldsymbol{S}_{i+1} - \boldsymbol{S}_i, \\
    \boldsymbol{M}_i^a & = \boldsymbol{M}_{i+1}^v - \boldsymbol{M}_i^v.
\end{aligned}
\label{equ-15}
\end{equation}
These dynamic representations encapsulate essential motion patterns. Next, for the given decoded features $\boldsymbol{D}_d$, we feed them into a fully connected layer to obtain the model-predicted target $\boldsymbol{M}^{pred}$.
\begin{equation}
\boldsymbol{M}^{pred} = FC(\boldsymbol{D}_d)
\label{equ-16}
\end{equation}
\noindent\textbf{Reconstruction Loss:} We use mean squared error (MSE) as the loss function. As shown in Fig.\ref{fig3}, joints contribute differently to motion representation, so the reconstruction loss must account for this variability. Instead of traditional binary masking, we introduce a weight-based masking strategy using normalized activation scores $\pi$ to assign adaptive weights to joints. This approach allows the loss function to prioritize the dynamic motion features, improving action representation learning.
\begin{equation}
\begin{aligned}
\mathcal{L}_v &= \frac{1}{|idx^{\text{mask}}|} \sum_{(i,j) \in idx^{\text{mask}}} \|\boldsymbol{M}_i^{\text{pred}} - \boldsymbol{M}_i^v\|_2^2 \cdot \pi_i, \\
\mathcal{L}_a &= \frac{1}{|idx^{\text{mask}}|} \sum_{(i,j) \in idx^{\text{mask}}} \|\boldsymbol{M}_i^{\text{pred}} - \boldsymbol{M}_i^a\|_2^2 \cdot \pi_i.
\end{aligned}
\label{equ-17}
\end{equation}
\hspace*{1em}The total reconstruction loss is then formulated as a convex combination of the velocity and acceleration losses: 
\begin{equation} \mathcal{L} = (1-\beta) \mathcal{L}_v + \beta \mathcal{L}_a \label{equ-18} 
\end{equation} 
Where $\beta$ is a hyperparameter controlling the relative contribution of the velocity and acceleration terms to the total loss.

\section{EXPERIMENTS}
\subsection{Datasets}
\noindent\textbf{NTU-RGB+D 60:} NTU-60 \cite{shahroudy2016ntu} is a large-scale dataset for skeleton-based action recognition with 56,880 3D skeleton sequences from 40 subjects across 60 action categories. It features two evaluation protocols: cross-subject (X-sub), with training on 20 subjects and testing on 20, and cross-view (X-view), with training on two camera views and testing on the third.

\noindent\textbf{NTU-RGB+D 120:} NTU-120 \cite{liu2019ntu} extends NTU-60 to 120 action categories, 114,480 skeleton sequences, and 106 subjects. It introduces the cross-setup (X-set) protocol, dividing 32 setups by camera distance and background into training and testing, while maintaining the cross-subject (X-sub) protocol. This provides a broader benchmark across subjects and environments.

\noindent\textbf{PKU Multi-Modality Dataset:} PKU-MMD \cite{liu2017pku} is a benchmark for human action recognition with around 20,000 samples across 51 action categories from 66 participants. It includes 1,076 video sequences captured from three viewpoints, split into two phases: PKU-MMD I (18,841 training, 2,704 testing) and PKU-MMD II (5,332 training, 1,613 testing), adding view variations and noise. Evaluations use a cross-subject protocol.

\subsection{Implementation Details}
\noindent\textbf{Network Architecture:} The MaskSem framework uses a transformer architecture with an Encoder and Decoder. The Encoder consists of $L_e$ = 8 identical blocks, while the Decoder has $L_d$ = 3 layers. Each block has an embedding dimension of 256, 8 multi-head self-attention modules, and a feed-forward network with a hidden dimension of 1024. Spatial and temporal positional embeddings are added before the first layer.\\
\noindent\textbf{Processing Skeletons:} During training, sequences are randomly cropped with a ratio $p$ in the range [0.5, 1], and resized to 120 frames. During pretraining, skeletons undergo random rotations and affine transformations for view diversity. A motion-aware masking ratio of 0.9 and a segment length of 4, as in MAMP \cite{mao2023masked}, are used.\\
\noindent\textbf{Pre-training stage:} A 90\% masking ratio is applied to the input tokens, with the AdamW optimizer, weight decay of 0.05, and betas (0.9, 0.95). Training lasts for 400 epochs with a batch size of 128. The learning rate increases linearly from 0 to 1e-3 during the first 20 warmup epochs, then decays to 5e-4 following a cosine schedule. Training is conducted in PyTorch on four NVIDIA RTX 4090 GPUs.

\begin{table*}[t!]
\caption{Performance comparison on the NTU-60, NTU-120, and PKU-MMD datasets under the linear evaluation protocol. * indicates that we retrain the models using their officially released code.
}\label{tab:tabel1}
\resizebox{\linewidth}{!}{
\centering
\begin{tabular}{cccccccccc}
\hline
\multirow{2}*{Method}&\multirow{2}*{Year}&\multirow{2}*{Input Stream}&\multicolumn{2}{c}{NTU-60}&\multicolumn{2}{c}{NTU-120}&PKU-I&PKU-II\\
&&&X-sub&X-view&X-sub&X-set&Phase I&Phase II\\
\hline
\multicolumn{9}{l}{\cellcolor{gray!20} \textit{Other pretext task:}} \\
LongTGAN\cite{zheng2018unsupervised}&AAAI'18&Joint&50.7&75.3&42.7&41.7&59.9&25.5 \\
P\&C\cite{su2020predict}&CVPR'20&Joint&50.7&75.3&42.7&41.7&59.9&25.5\\
\hline
\multicolumn{9}{l}{\cellcolor{gray!20} \textit{Contrastive learning:}} \\

CrosSCLR\cite{li20213d}&CVPR'21&Joint+Motion+Bone&77.8&83.4&67.9&66.7&84.9&21.2\\
PSTL\cite{zhou2023self}&AAAI'23&Joint+Motion+Bone&79.1&83.8&69.2&70.3&89.2&52.3\\
ActCLR\cite{lin2023actionlet}&CVPR'23&Joint+Motion+Bone&84.3&88.8&74.3&75.7&-&-\\
HiCo-former\cite{dong2023hierarchical}&AAAI'23&Joint&81.1&88.6&72.8&74.1&89.3&49.4\\
{H$^2$E \cite{chen2023self}} & TIP'23 & Joint & 78.7 & 82.3 & - & - & 88.5 & 51.7 \\
{PCM$^3$ \cite{zhang2023prompted}} & MM'23 & Joint & 83.9 & \underline{90.4} & 76.5 & 77.5 & - & 51.5 \\
Skeleton-logoCLR\cite{hu2024global}&TCSVT'24&Joint&82.4&87.2&72.8&73.5&90.8&54.7\\
\hline
\multicolumn{9}{l}{\cellcolor{gray!20} \textit{Masked prediction:}} \\

SkeletonMAE\cite{wu2023skeletonmae}&ICME’23&Joint&74.8&77.7&72.5&73.5&82.8&36.1\\
MAMP$^*$\cite{mao2023masked}&ICCV’23&Joint&84.9&89.1&\underline{78.2}&79.1&91.6&53.8\\
MMFR\cite{zhu2024motion}&TCSVT’24&Joint&84.2&89.5&77.1&78.8&\textbf{92.4}&\underline{54.4}\\
S-JEPA\cite{abdelfattah2024s}&ECCV’24&Joint&\underline{85.3}&89.8&\textbf{79.6}&\textbf{79.9}&\underline{92.2}&53.5\\
\textbf{MaskSem(Ours)} &-&Joint&\textbf{85.9}&\textbf{90.8}&77.5&\underline{79.3}&91.2&\textbf{55.8}\\
\hline
\end{tabular}
}

\end{table*}

\begin{table*}[!ht]
\caption{Performance comparison on the NTU-60 and NTU-120 datasets under the fine-tuning evaluation protocol. * indicates that we retrain the models using their officially released code.
}\label{tab:tabel2}
\centering
\resizebox{\linewidth}{!}{
\begin{tabular}{cccccccccc}
\hline
\multirow{2}*{Method}&\multirow{2}*{Year}&\multirow{2}*{Input Stream}&\multirow{2}*{Backbone}&\multicolumn{2}{c}{NTU-60}&\multicolumn{2}{c}{NTU-120}\\
&&&&X-sub&X-view&X-sub&X-set\\
\hline
\multicolumn{8}{l}{\cellcolor{gray!20} \textit{Other pretext task:}} \\
Colorization\cite{yang2021skeleton}&ICCV'21&Joint+Motion+Bone&DG-CNN&88.0&94.9&-&-\\
Hi-TRS\cite{chen2022hierarchically}&ECCV'22&Joint+Motion+Bone&Transformer&90.0&95.7&85.3&87.4\\
\hline
\multicolumn{8}{l}{\cellcolor{gray!20} \textit{Contrastive learning:}} \\
CrosSCLR\cite{li20213d}&CVPR'21&Joint+Motion+Bone&ST-GCN&86.2&92.5&80.5&80.4\\
HYSP\cite{guo2022contrastive}&AAAI'22&Joint+Motion+Bone&ST-GCN&89.1&95.2&84.5&86.3\\
ActCLR\cite{lin2023actionlet}&CVPR'23&Joint+Motion+Bone&ST-GCN&88.2&93.9&82.1&84.6\\
Skeleton-logoCLR\cite{hu2024global}&TCSVT'24&Joint+Motion+Bone&ST-GCN&89.4&94.3&84.6&85.7\\
\hline
\multicolumn{8}{l}{\cellcolor{gray!20} \textit{Masked prediction:}} \\
SkeletonMAE\cite{wu2023skeletonmae}&ICME’23&Joint&STTFormer&86.6&92.9&76.8&79.1\\
SkeletonMAE\cite{wu2023skeletonmae}&ICME’23&Joint&Transformer&88.5&94.7&87.0&88.9\\
SkeletonMAE\cite{yan2023skeletonmae}&ICCV’23&Joint&STRL&92.8&96.5&84.8&85.7\\
MotionBERT\cite{zhu2023motionbert}&ICCV’23&Joint&DSTformer&\underline{93.0}&97.2&-&-\\
MAMP$^*$\cite{mao2023masked}&ICCV’23&Joint&Transformer&92.7&97.5&\underline{89.7}&\textbf{91.3}\\
MMFR\cite{zhu2024motion}&TCSVT’24&Joint&Transformer&91.9&96.5&87.4&\underline{90.4}\\
S-JEPA\cite{abdelfattah2024s}&ECCV’24&Joint&Transformer&\textbf{93.1}&\underline{97.6}&\textbf{90.3}&\textbf{91.3}\\
\textbf{MaskSem(Ours)} &-
&Joint&Transformer&\underline{93.0}&\textbf{97.7}&89.4&\textbf{91.3}\\
\hline
\end{tabular}
}
\end{table*}

\subsection{Comparison with State-of-the-art Methods}
\noindent\textbf{Linear Evaluation:} We fix the pre-trained backbone and train a linear classifier for 100 epochs with a batch size of 256 and an initial learning rate of 0.1, which decays via a cosine schedule. Performance on the NTU-60, NTU-120 datasets (Table \ref{tab:tabel1}) shows MaskSem improves over PCM$^3$ \cite{zhang2023prompted} by 2.0\% and 1.6\% on NTU-60 X-Sub \cite{shahroudy2016ntu} and NTU-120 X-Sub \cite{liu2019ntu}, respectively, and by 1.0\% and 1.8\% on X-Set. MaskSem outperforms MAMP \cite{mao2023masked} on four out of six dataset subsets, demonstrating that the model has learned more discriminative features relevant to downstream tasks from our approach. However, MaskSem shows performance degradation on the NTU-120 and PKU-I X-Sub, which may be due to the inability to fully capture relative motion information from a single perspective, leading to poor generalization on the multi-class Sub of the dataset.

\noindent\textbf{Fine-tuning Evaluation:} In the fine-tuning protocol, an MLP head is added to the pre-trained backbone, and the entire network is fine-tuned for 100 epochs with a batch size of 48. The learning rate is linearly increased from 0 to 3e-4 during the first 5 warm-up epochs, then decayed to 1e-5 using a cosine schedule, with layer-wise learning rate decay as in \cite{bao2021beit}. Fine-tuned performance on the NTU-60 and NTU-120 datasets (Tab. \ref{tab:tabel2}) shows that MaskSem outperforms all existing contrastive learning methods, leading by 3.6\%, 4.8\%, 5.4\%, and 5.6\% on four subsets.

\begin{table}[!h]
\caption{Accuracy comparison under the semi-supervised evaluation protocol.\label{tab:table3}}
\centering
\resizebox{0.5\textwidth}{!}{
\begin{tabular}{cccccccccc}
\hline
\multirow{3}*{Method}&\multirow{3}*{Year}&\multicolumn{4}{c}{NTU-60}\\
&&\multicolumn{2}{c}{X-sub}&\multicolumn{2}{c}{X-view}\\
&&(1\%)&(10\%)&(1\%)&(10\%)\\
\hline
\multicolumn{6}{l}{\cellcolor{gray!20} \textit{Other eetext task:}} \\
Colorization\cite{yang2021skeleton}&ICCV'21&48.3&71.7&52.5&78.9\\
Hi-TRS\cite{chen2022hierarchically}&ECCV'22&49.3&77.7&51.5&81.1\\
\hline
\multicolumn{6}{l}{\cellcolor{gray!20} \textit{Contrastive learning:}} \\
3s-CrosSCLR\cite{li20213d}&CVPR'21&51.1&74.4&50.0&77.8\\
3s-CMD\cite{hua2023part}&ECCV’22&55.6&79.0&55.5&82.4\\
3s-HYSP\cite{guo2022contrastive}&ICLR'23&-&80.5&-&85.4\\
3s-SkeAttnCLR\cite{hua2023part}&IJCAI’23&59.6&81.5&59.2&83.8\\
\hline
\multicolumn{6}{l}{\cellcolor{gray!20} \textit{Masked prediction:}} \\
SkeletonMAE\cite{wu2023skeletonmae}&ICME’23&54.4&80.6&54.6&83.5\\
MAMP\cite{mao2023masked}&ICCV’23&66.0&\underline{88.0}&68.7&91.5\\
MMFR\cite{zhu2024motion}&TCSVT’24&65.0&87.0&\underline{71.3}&91.0\\
S-JEPA\cite{abdelfattah2024s}&ECCV’24&\underline{67.5}&\textbf{88.4}&69.1&\underline{91.5}\\
\textbf{MaskSem(Ours)}&-&\textbf{68.4}&\textbf{88.4}&\textbf{72.5}&\textbf{92.2}\\
\hline

\end{tabular}
}
\end{table}

\noindent\textbf{Semi-supervised Evaluation:} In the semi-supervised protocol, we fine-tune the post-attached classification layer and pre-trained Encoder using a small portion of the labeled training data, with other settings matching the fine-tuned evaluation protocol. As in \cite{li20213d,guo2022contrastive}, performance is evaluated on the NTU-60 dataset with 1\% and 10\% of labeled training data, and the average of five runs is reported to mitigate randomness. As shown in Tab. \ref{tab:table3}, MaskSem achieves state-of-the-art performance, outperforming the baseline MAMP by 2.4\%, 0.4\%, 3.8\%, and 0.7\% across four subsets, confirming its efficiency in semi-supervised scenarios.

\noindent\textbf{Transfer Learning Evaluation Results:} We pre-trained MaskSem on the NTU-60 X-Sub and NTU-120 X-Sub datasets and then fine-tuned it on the target dataset PKU-MMD II. As shown in Table \ref{tab:table4}, transferring the model from NTU-60 to PKU-MMD II resulted in the best performance, highlighting MaskSem's strong transferability. However, there is a performance degradation on NTU-120, which may be due to overfitting, limiting the model's generalization ability.
\begin{table}[!h]
\caption{Performance comparison under the transfer learning protocol.\label{tab:table4}}
\centering
\resizebox{0.5\textwidth}{!}{
\begin{tabular}{cccccccccc}
\hline
\multirow{2}*{Method}&\multirow{2}*{Journal\&Year}&\multicolumn{2}{c}{To PKU-II}\\
&&NTU-60&NTU-120\\
\hline
LongT GAN\cite{zheng2018unsupervised}&AAAI'18&44.8&-\\
ISC\cite{thoker2021skeleton}&MM'21&45.8&-\\
SkeletonMAE\cite{yan2023skeletonmae}&ICME'21&58.4&61.0\\
CMD\cite{mao2022cmd}&ECCV'22&56.0&57.0\\
MAMP\cite{mao2023masked}&ICCV'23&70.6&\underline{73.2}\\
MMFR\cite{zhu2024motion}&TCSVT'24&68.7&69.7\\
S-JEPA\cite{abdelfattah2024s}&ECCV'24&\underline{71.4}&\textbf{74.2}\\
\textbf{MaskSem(ours)}&-&\textbf{71.6}&72.2\\
\hline
\end{tabular}
}
\end{table}
\subsection{Ablation Study}
Extensive ablation experiments were performed on the NTU-60 X-view dataset to evaluate the proposed MaskSem. Unless stated otherwise, all results are presented using the linear evaluation protocol.

\begin{table}[!h]
\caption{ABLATION STUDY ON THE MASKING STRATEGY AND RECONSTRUCTION TARGET.\label{tab:table5}}
\centering
\resizebox{0.5\textwidth}{!}{
\begin{tabular}{cccccc}
\hline
\multicolumn{2}{c}{Masking}&\multicolumn{2}{c}{Target}&\multirow{2}*{Acc(\%}\\
motion&semantic&Velocity&Acceleration\\
\hline
\checkmark&&\checkmark&&89.1\\
&\checkmark&\checkmark&&90.3\\
\checkmark&&\checkmark&\checkmark&90.0\\
&\checkmark&&\checkmark&85.8\\
&\checkmark&\checkmark&\checkmark&\textbf{90.8}\\
\hline
\end{tabular}
}
\end{table}

\noindent\textbf{Mask Strategy and Reconstruction Target:} To validate the effectiveness of our proposed method, we compared it with the baseline, as shown in Table \ref{tab:table5}. Our proposed semantic-guided masking approach outperforms motion-aware masking by 1.2\%. This indicates that the semantic information of joints, as prior knowledge, can more effectively guide the skeleton masking process. Our proposed hybrid high-order motion as the reconstruction target improves performance by 0.9\% compared to using only low-order motion. This demonstrates that multi-order motion can reduce the impact of noise and help the model learn richer motion information. However, if only high-order motion is used as the reconstruction target, performance decreases by 3.3\%, indicating that first-order differences contain richer motion information than second-order differences.

\begin{figure}[h!]
    \centering
    \includegraphics[width=0.8 \linewidth,trim=40 10 80 60,clip]{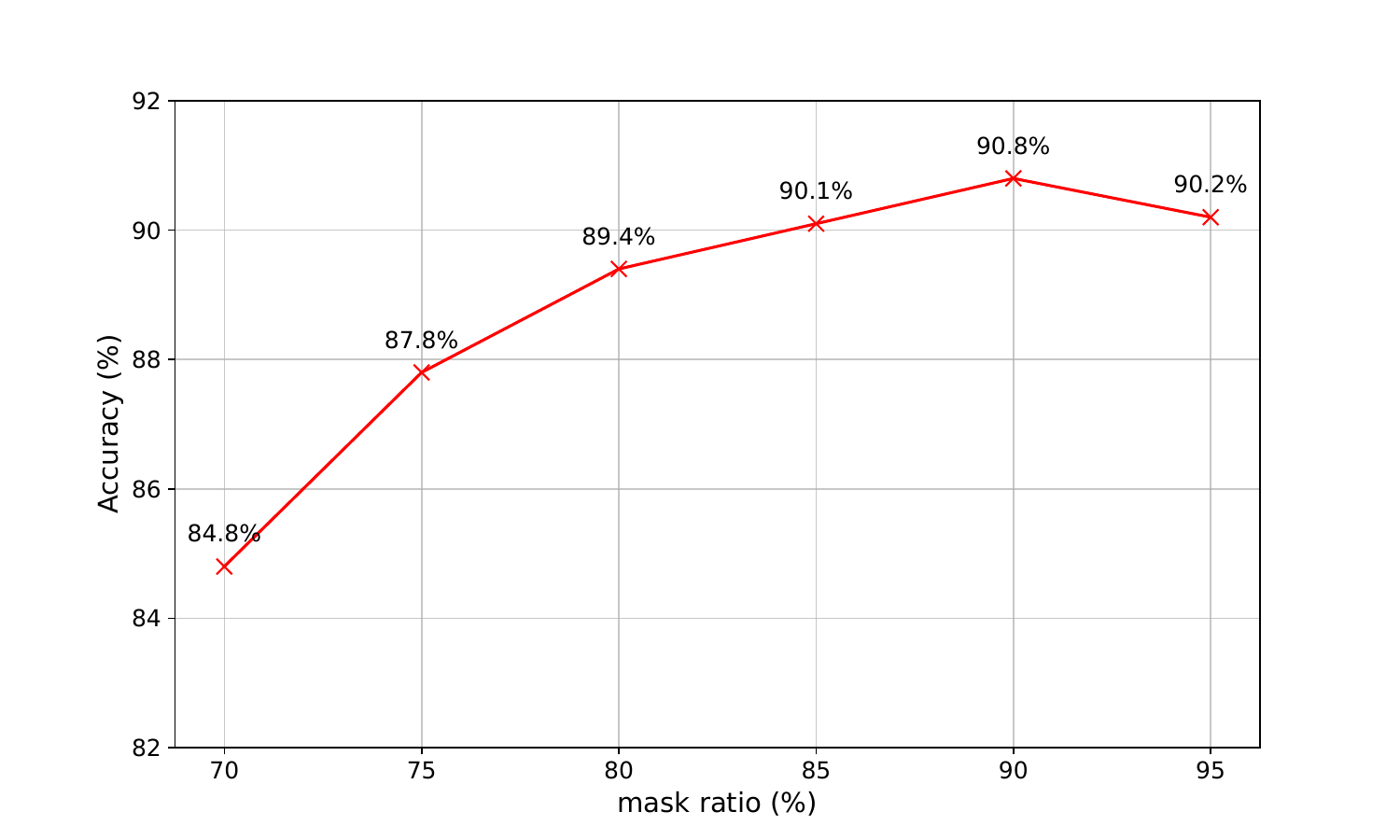}
    \caption{Ablation study on the masking ratio on NTU-60 X-View dataset.}
    \label{fig4}
\end{figure}

\noindent\textbf{Masking Ratio:} As illustrated in Figure \ref{fig4}, we explored various masking ratios. Results on the NTU-60 X-view dataset indicate that both excessively large and small masking ratios negatively affect performance. Empirically, we observe that a 90\% masking ratio yields the best results.

\begin{figure}[h!]
    \centering
    \includegraphics[width=0.8 \linewidth,trim=40 10 80 60,clip]{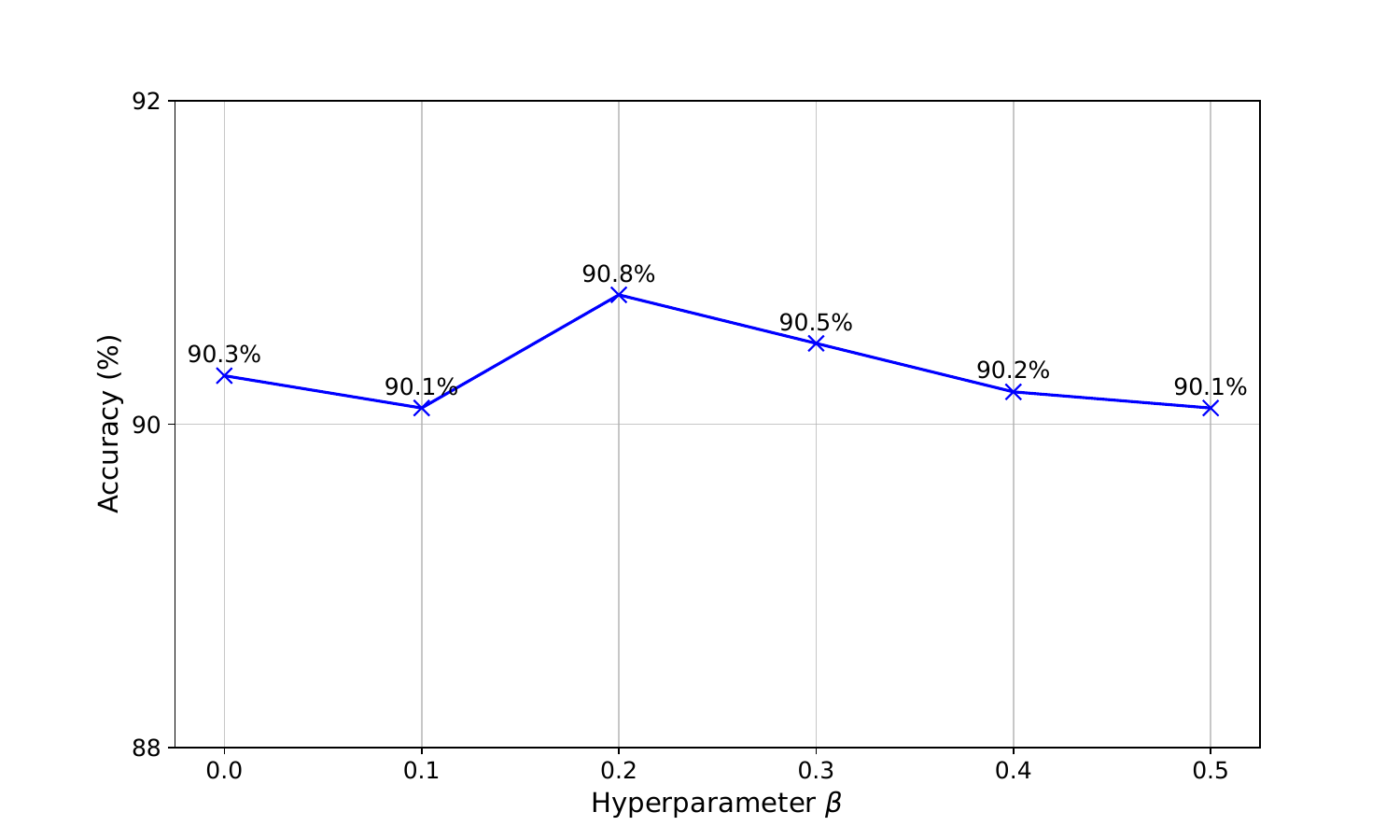}
    \caption{Ablation study on the Hyperparameter $\boldsymbol{\beta}$ on NTU-60 X-View dataset.}
    \label{fig5}
\end{figure}

\noindent\textbf{Hyperparameter $\boldsymbol{\beta}$:} As illustrated in Figure \ref{fig5}, the ablation study reveals that $\beta = 0.2$ achieves the best performance. This indicates that emphasizing velocity ($\mathcal{L}_v$) while giving moderate attention to acceleration ($\mathcal{L}_a$) 
captures the most critical motion features, ensuring smooth and realistic motion representation without overfitting to high-frequency dynamics.

\section{CONCLUSIONS}
This paper introduces MaskSem, a self-supervised framework for skeleton-based action recognition aimed at enhancing robotic systems. By combining semantic-guided dynamic joint masking with hybrid high-order motion reconstruction, MaskSem addresses key limitations of existing methods. The semantic-guided masking prioritizes the most important joints and encourages the model to explore joints with more discriminative information. Hybrid high-order motion reconstruction captures subtle motion details from multi-order motion patterns. Experiments on the NTU60, NTU120, and PKU-MMD datasets highlight MaskSem’s effectiveness, making it well-suited for robotic action recognition in interactive scenarios.

\section*{Acknowledgment}
This work was supported by the National Natural Science Foundation of China (Grant No. 62173045), the Beijing Natural Science Foundation (Grant No. F2024203115), the China Postdoctoral Science Foundation (Grant No. 2024M750255), the National Key R\&D Program of China (Grant No. 2024YFC3015604), and the ``Double First-Class'' Interdisciplinary Team Project of Beijing University of Posts and Telecommunications (Grant No. 2023SYLTD02).

\bibliographystyle{IEEEtran}
\bibliography{bib}
\end{document}